\documentclass[journal]{IEEEtran}

\usepackage{cite}
\usepackage[final]{pdfpages}
\usepackage[vlined, ruled, linesnumbered, commentsnumbered]{algorithm2e}
\usepackage[cmex10]{amsmath}
\usepackage{color}
\definecolor{chred}{rgb}{0.8,0,0}
\definecolor{chblue}{rgb}{0,0,0.7}
\definecolor{chgray}{rgb}{0.5,0.5,0.5}
\usepackage{graphicx}
\usepackage{caption}
\usepackage{dblfloatfix}
\usepackage{subfigure}
\usepackage{eqlist}
\usepackage{txfonts}
\usepackage{lipsum}
\usepackage{url}
\usepackage{footmisc}
\usepackage{booktabs}

\usepackage{multirow}
\usepackage{threeparttable}
\usepackage{soul}
\newcommand{\gr}[1]{\textcolor{gray}{#1}}

\newcommand{\rev}[1]{\textcolor{black}{#1}}

\def\HiLi{\leavevmode\rlap{\hbox to \hsize{\color{yellow!50}\leaders\hrule height .8\baselineskip depth .5ex\hfill}}}

\begin{document}

\title{Preparatory Manipulation Planning using Automatically Determined
Single and Dual Arms}

\author{Weiwei~Wan,
        Kensuke~Harada, and Fumio~Kanehiro
\thanks{W. Wan and K. Harada are with Osaka University, Japan.
F. Kanehiro is with National Inst. of AIST, Japan.
{\tt\small wan-weiwei@aist.go.jp}}}

\markboth{Journal of \LaTeX\ Class Files,~Vol.~x, No.~x, xxxx~yyyy}%
{Shell \MakeLowercase{\textit{et al.}}: Bare Demo of IEEEtran.cls for IEEE Journals}

\maketitle

\begin{abstract}
This paper presents a manipulation planning algorithm for robots to reorient
objects. It automatically finds a sequence of robot motion that manipulates and
prepares an object for specific tasks.
Examples of the preparatory manipulation planning problems include reorienting
an electric drill to cut holes, reorienting workpieces for assembly, and
reorienting cargo for packing, etc. The proposed algorithm could plan single and
dual arm manipulation sequences to solve the problems. The mechanism under the
planner is a regrasp graph which encodes grasp configurations and object poses. The
algorithms search the graph to find a sequence of robot motion to reorient
objects. The planner is able to plan both single and dual arm manipulation. It
could also automatically determine whether to use a single arm, dual arms, or their
combinations to finish given tasks. The planner is examined by various humanoid
robots like Nextage, HRP2Kai, HRP5P, etc., using both simulation and real-world
experiments.
\end{abstract}

\begin{IEEEkeywords}
Manipulation Planning, Reorienting Objects, Dual-arm Robots
\end{IEEEkeywords}

\IEEEpeerreviewmaketitle

\section{Introduction}\label{sec:introduction}

\begin{figure}[!htbp]
	\centering
	\includegraphics[width=.487\textwidth]{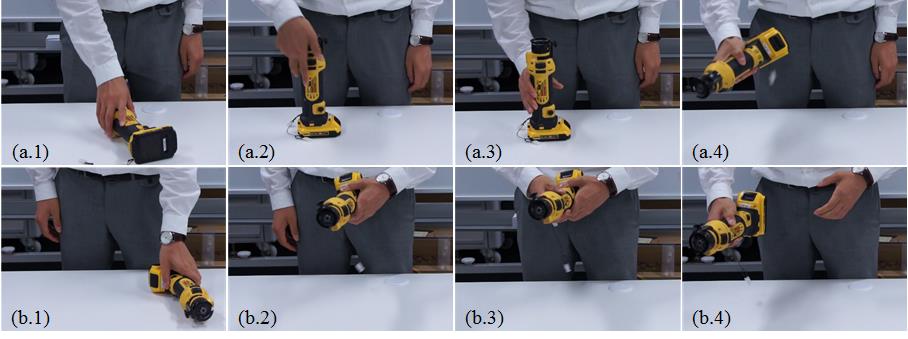}
	\caption{(a.1-a.4): Preparatory manipulation of an electric drill using a
	single arm. (b.1-b.4): The same preparatory manipulation using dual-arm
	handover.}
	\label{teaser}
\end{figure}

\IEEEPARstart{P}{reparatory} manipulation planning automatically finds a
sequence of robot motion that manipulates and prepares an object for specific
tasks \cite{Nguyen16}. The reason preparatory manipulation is needed is that
objects in the real world are not always well posed. Take Fig.\ref{teaser}
for example. In the figure, an electric drill is in two initial poses shown in
Fig.\ref{teaser}(a.1) and Fig.\ref{teaser}(b.1).
To use the drill, a human worker has to reorient the drill to
let the \rev{tooltip} face to targets like Fig.\ref{teaser}(a.4)
and (b.4). The reorienting process is named preparatory manipulation.

Fig.\ref{teaser} shows two ways to do preparatory manipulation.
The first one is single-arm regrasp shown in Fig.\ref{teaser}(a.1-a.4), where
the human worker picks up the electric tool using his right hand in (a.1),
places it down on the table in (a.2), changes the pose of his right hand to
regrasp the tool in (a.3), and successfully reorients the \rev{tooltip} in
(a.4). Only a single arm (the right arm and right hand) is used in the
process. The second one is dual-arm handover shown in
Fig.\ref{teaser}(b.1-b.4), where the human picks up the electric tool using his
left hand in (b.1), moves it to a hand-over pose in (b.2), hands it over to his
right hand in (b.3), and reorients the \rev{tooltip} in (b.4).
Both arms are used in the second case.

Preparing objects using single-arm and dual-arm manipulation is
common and important to handle objects in our daily life. It motivates us to
develop algorithms for robots to have similar manipulation skills. In the
paper, we challenge the difficulty of planning multiple robot
motions together and automatically selecting the necessary number of arms. We develop
algorithms to find a sequence of collision-free and IK-feasible motion to move
and reorient objects, and propose a super regrasp graph that helps to determine
the sequence of multi-arm handover and whether to use a single arm, dual arms,
or their combinations to perform given tasks.
Lots of related studies have been devoted to the topic in the past few decades.
They are summarized in the related work section. Compared with them, this paper
initially performs preparatory manipulation planning using fully automatically
generated grasps and object poses. It is also the first work that
automatically determines the choices of arms.

The mechanism under the planner is a regrasp graph, where each node in the graph
represents one grasp configuration.
Given the initial and goal poses of an object, the planner connects the
accessible grasps of initial and goal poses of an object to the graph and
searches the graph to find a sequence of pick-and-place-based preparatory
manipulation motion. For dual-arm manipulation, the planner builds two regrasp
graphs where each one encodes the grasps of one hand. The planner connects the
two graphs at hand-over poses to make a super regrasp graph, and searches the
super regrasp graph to find a sequence of preparatory manipulation motion. The
results could be either pick-and-place using a single arm, hand-over using dual arms, or mixed
single and dual-arm operations, depending on feasibility and lengths of paths
found by the graph search algorithms.

The planner is part of an open source project named PYHIRO\footnote{
Project link is excluded for review.}. It is
examined by various humanoid robots like HRP2Kai, Nextage, etc, using both
simulation and real-world experiments.

\section{Related Work and Contributions}

There are two problems in preparatory manipulation planning.
One is to find a sequence of collision-free and IK-feasible motion to
move and reorient objects. The problem, in this case, is usually named the
sequential manipulation planning problem or the combined task and motion planning problem.
The other one is to incorporate multiple arms and determines the sequence of
multi-arm handover and whether to use a single arm or dual arms.

Lots of studies have been devoted to the first problem. Some early publications
include \cite{Pierre87}\cite{Hajime98},
which used grasp-placement tables to find a sequence of preparatory motion. The
grasp-placement tables orchestrate the stable states and accessible grasps of an
object to provide a data structure for searching manipulation sequences. The
grasps in these early work were predefined or generated by planners designed for specific
grippers, and the objects were simple polygonal objects.

Recent work included more complicated grasp planners and was integrally
considered with backtracking, geometrical reasoning, and motion planning. For
example, Simeon et al. \cite{Thierry04} presented a framework which integrated motion planning
and preparatory manipulation planning to move objects among obstacles. Xue et
al. \cite{Zhixing08} used shape primitives 
to plan the grasps of
a  multi-finger hand and implemented the preparatory manipulation planning of a cup
using the multi-finger hand. Bohg et al. \cite{Bohg12} used Bayesian Network to
model task constraints and plan the preparatory manipulation sequence of
containers and toys. King et al.
\cite{King2013} used integral primitive-based prehensile and non-prehensile
grasp planning to exercise preparatory manipulation of everyday objects.
Lozano-Perez et al. \cite{Tlp14} used symbolic search to find manipulation
sequences under various constraints. The method was used to plan sequences for a mobile
manipulator to rearrange objects. Similarly,
Srivastava et al. \cite{Srivastava14}, Krontiris et al. \cite{Krontiris15},
Toussaint \cite{Toussaint15}, 
and Dantam et al. \cite{Dantam16} respectively
used symbolic reasoning to remove obstacles and pick up or rearrange objects.
Especially, Toussaint \cite{Toussaint15} and Dantam et al. \cite{Dantam16}
presented algorithms to plan motion sequences for robots to stack
distributed objects, considering both task and motion constraints.
Mirabel et al. \cite{mirabel2017manipulation} presented a planner
considering the constraints from coordinated manipulators.
More extensively, Lee et al. \cite{Lee15} and Woodruff et al.
\cite{woodruff2017planning} respectively presented a framework work that used
integral prehensile and non-prehensile planning to generate sequential
manipulation. 


There are also lots of studies on dual-arm or multi-arm preparatory
manipulation planning. For example, Koga et al.
\cite{Yoshihito94b} used predefined grasps
to plan sequential manipulation for multiple manipulators in early time. Cho
et al. \cite{Cho03} applied the grasp-placement tables developed by
\cite{Pierre87} to dual-arm robots. Saut et al. \cite{Jean10} used
decomposition to plan grasps and implemented dual-arm regrasp of complicated
mesh models. Graphs, instead of grasp-placement tables, were used in the
planner. Harada et al. \cite{Harada14b} presented regrasp and hand-over
planning of dual-arm robots across multi-modal configuration spaces, and implemented a
practical system in \cite{Harada14}. Details of multi-modal
motion planning could be found in \cite{Kris10}, which made a concrete
description of the concept and presented several implementations.
More recent work like \cite{Hayashi17} used dual-arm sequential manipulation
planning to wrap up the fabric. It is also quite relevant except that
reorienting directions are constrained by the wrapping heuristics. Lertkultanon et al.
\cite{Lertkultanon17}, Zhou et al. \cite{xian2017closed}, and Suarez-Rui
et al. \cite{suarez2018can} respectively applied integral prehensile and
non-prehensile sequential dual-arm motion planning to the preparatory manipulation of a chair.
Vezzani et al. \cite{vezzani2017novel} presented a framework for
simultaneous handover and visual recognition.

Compared with the related work, our contribution is a super regrasp graph which
is extensible over multi-modal configuration spaces and multiple arms.
The super regrasp graph is composed of components like \textit{partial regrasp
graphs for the initial and goal poses}, \textit{single-arm regrasp graphs}, and
\textit{hand-over regrasp graphs}. The proposed planner could search the super
regrasp graph to automatically decide whether to use a
single arm or both arms to finish given tasks. To our best knowledge, this paper
is the first work that automatically chooses the number of arms and
uses mixed single and dual-arm operations for preparatory manipulation.

\section{Multi-modal Motion Planning}

\subsection{The theory}

The fundamental theory supporting preparatory manipulation planning is
multi-modal motion planning \cite{Kris10}, which means
to plan paths across multiple configuration spaces, and output a sequence of transfer and transit motion.

\begin{figure}[!htbp]
	\centering
	\includegraphics[width=.487\textwidth]{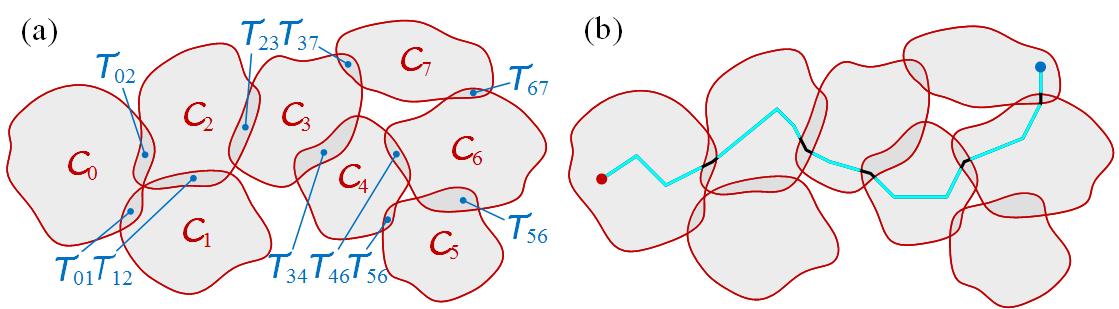}
	\caption{An example of multi-modal motion planning. (a) Planning across seven
	configuration spaces. (b) A path found by the multi-modal planner.}
	\label{multimodal}
\end{figure}

Fig.\ref{multimodal} sketches an example of multi-modal motion planning. Here,
the goal is to plan robot motion across eight configuration spaces named
$\mathcal{C}_0$, $\mathcal{C}_1$, \ldots, $\mathcal{C}_7$. A configuration
space is defined as:
\begin{equation}
	\mathcal{C}=\{d_g, q_0, q_1, \ldots, q_n \in T^{n+2}\}
\end{equation}
where $T^{n+2}$ indicates an n+2-dimensional topology torus. $d_g$ denotes the
jawwidth of a robotic gripper, $q_i$ indicates the $i$th joint angle of a
manipulator.
The seven configuration spaces are connected to each other
by overlapped manifolds.
As is shown in Fig.\ref{multimodal}(a), the areas named $\mathcal{T}_{ij}$ are
the overlapped manifolds.
A $\mathcal{T}_{ij}$ connects two adjacent configuration spaces
$\mathcal{C}_i$ and $\mathcal{C}_j$. It is composed by the shared configurations of
$\mathcal{C}_i$ and $\mathcal{C}_j$. Formal definitions of $\mathcal{T}_{ij}$
will be presented in the next subsection. In multi-modal motion planning,
the overlapped manifolds are named transit spaces. The remaining configuration
spaces are named transfer spaces. Fig.\ref{multimodal}(b) shows a path found by the multi-modal planner.
The initial configuration is denoted by a red point. It is in configuration
space $\mathcal{C}_0$. The goal configuration is denoted by a blue point. It is in
configuration space $\mathcal{C}_7$. The planner finds a path crossing
six configuration spaces $\mathcal{C}_0$, $\mathcal{C}_2$,
$\mathcal{C}_3$,  $\mathcal{C}_4$, $\mathcal{C}_6$, and $\mathcal{C}_7$. Along
the path, the black edges connect two identical poses, they are transfer
edges. The cyan edges connect two different poses, they are the transit edges.
The planned result is a sequence of transfer and transit motion along the transfer
and transit edges.

\subsection{The modalities of preparatory manipulation}

\begin{figure}[!htbp]
	\centering
	\includegraphics[width=.487\textwidth]{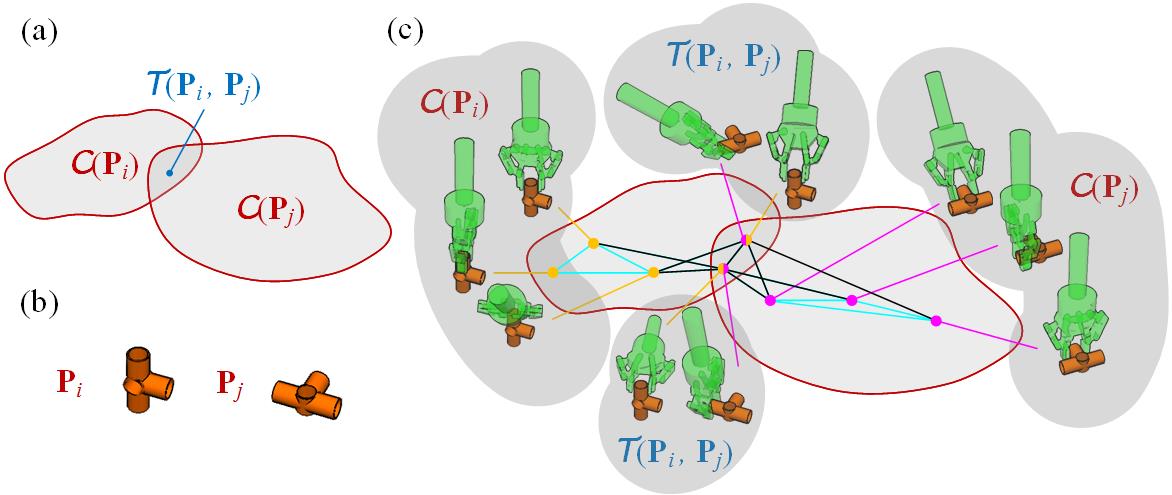}
	\caption{(a) Two modals. $\mathcal{C}$($\mathbf{P}^i$) and
	$\mathcal{C}$($\mathbf{P}^j$) are the transfer spaces.
	$\mathcal{T}$($\mathbf{P}^i$, $\mathbf{P}^j$) is the transit manifold. (b) The
	two object poses identifying the two modals. (c) Sampled nodes in
	$\mathcal{C}$($\mathbf{P}^i$), $\mathcal{C}$($\mathbf{P}^j$), and $\mathcal{T}$($\mathbf{P}^i$,
	$\mathbf{P}^j$). Each node corresponds to one collision-free and IK-feasible
	grasp.}
	\label{multimodalmp}
\end{figure}

Solving a multi-modal motion planning problem requires properly identifying
the modalities. Considering that the goal of the preparatory manipulation
planning is (1) to reorient the poses of objects, and (2) to find a sequence
of arm motion, we use the configuration space of a robot arm as a transfer
space, and use object poses to identify various modals.
Fig.\ref{multimodalmp}(a) exemplifies the identification. Each modal is
essentially the configuration space of a single arm. It is varied by object
poses and is written as $\mathcal{C}$($\mathbf{P}^i$), which means the
configuration space is modalized by an object pose $\mathbf{P}^i$. 
Let $\mathbf{P}^i$ be defined as:
\begin{equation}
	\mathbf{P}^i=\{p_x^i, p_y^i, p_z^i, r_x^i, r_y^i, r_z^i \in SE(3)\}
\end{equation}
$p_x^i$, $p_y^i$, $p_z^i$, $r_x^i$, $r_y^i$, $r_z^i$ denote the position and
orientation components of the object pose.
A modalized configuration
space could be written as:
\begin{equation}
	\mathcal{C}(\mathbf{P}^i) = \{d_g, q_0, q_1, \ldots, q_n, p_x^i, p_y^i,
	p_z^i, r_x^i, r_y^i, r_z^i \in T^{n+2}\times SE(3)\}
	\label{posdimfull}
\end{equation}
The jawwidth $d_g$ is a value irrelevant to the pose of an object.
The total dimension of a modalized configuration space is $T^{n+2}\times SE(3)$.
It is a n+8-dimensional manifold.
By replacing the items using 
\begin{equation}
	\textbf{q}=\{q_0, q_1, \ldots, q_n\}
\end{equation}
and
\begin{equation}
	\textnormal{\textbf{T}}_{\textnormal{\textbf{P}}^i}=
	\begin{bmatrix}
	c_{r_y^i}c_{r_z^i} & -c_{r_y^i}s_{r_z^i} & s_{r_y^i} & p_x^i\\
	s_{r_x^i}s_{r_y^i}c_{r_z^i}+c_{r_x^i}s_{r_z^i} &
	-s_{r_x^i}s_{r_y^i}s_{r_z^i}+c_{r_x^i}c_{r_z^i} & -s_{r_x^i}c_{r_y^i} & p_y^i\\
	-c_{r_x^i}s_{r_y^i}c_{r_z^i}+s_{r_x^i}s_{r_z^i} &
	c_{r_x^i}s_{r_y^i}s_{r_z^i}+s_{r_x^i}c_{r_z^i} & c_{r_x^i}c_{r_y^i} & p_z^i\\
	0 & 0 & 0 & 1\\
	\end{bmatrix}
\end{equation}
, where
$s_{r_x^i}$ = $sin(r_x^i)$, $c_{r_x^i}$ = $cos(r_x^i)$, $s_{r_y^i}$ =
$sin(r_y^i)$, $c_{r_y^i}$ = $cos(r_y^i)$, $s_{r_z^i}$ = $sin(r_z^i)$, $c_{r_z^i}$
= $cos(r_z^i)$, equation (\ref{posdimfull}) can be rewritten as:
\begin{equation}
	\mathcal{C}(\mathbf{P}^i) = \{d_g, \textbf{q},
	\textnormal{\textbf{T}}_{\textnormal{\textbf{P}}^i}\}
	\label{posdim}
\end{equation}
\indent
Two modals,
for example $\mathcal{C}$($\mathbf{P}^i$) and $\mathcal{C}$($\mathbf{P}^j$)
shown in the Fig.\ref{multimodalmp}(a), are connected to each other by a
transit manifold named $\mathcal{T}$($\mathbf{P}^i$, $\mathbf{P}^j$). The
transit manifold could be either modalized by $\mathbf{P}^i$ or $\mathbf{P}^j$.
When modalized by $\mathbf{P}^i$, the manifold
is defined as $\{d_g, \textbf{q},
\textnormal{\textbf{T}}_{\textnormal{\textbf{P}}^i}\}$.
When modalized by $\mathbf{P}^j$, the manifold is defined as
$\{d_g, \textbf{q}, \textnormal{\textbf{T}}_{\textnormal{\textbf{P}}^j\}}$
The two definitions are equal to each in the object's local coordinate system:
\begin{equation}
	\textnormal{\textbf{T}}_{\textnormal{\textbf{P}}^i}\cdot
	J(\textbf{q}^{(i)}) = \textnormal{\textbf{T}}_{\textnormal{\textbf{P}}^j}\cdot
	J(\textbf{q}^{(j)})
	\label{posequal}
\end{equation}
Here, $J(\textbf{q}^{(i)})$ and $J(\textbf{q}^{(j)})$ indicate the Jacobian
matrix of a robot under joint angles $\textbf{q}$ from configuration spaces
modalized by $\mathbf{P}^i$ and $\mathbf{P}^j$ respectively.

Equations (\ref{posdimfull}-\ref{posequal}) show: (1) A configuration in
$\mathcal{T}$($\mathbf{P}_i$, $\mathbf{P}_j$) is on a manifold; (2) The configuration belongs to both
$\mathcal{C}$($\mathbf{P}^i$) and $\mathcal{C}$($\mathbf{P}^j$);
(3) The configuration implies a grasp that is identical in the local coordinate
system of the object.

Fig.\ref{multimodalmp}(b) shows the configurations and their relations in
the modalized configuration spaces. The configuration spaces are sampled. Each
node indicates a sampled configuration.
Especially, the two nodes
under $\mathcal{T}$($\mathbf{P}_i$, $\mathbf{P}_j$) correspond to four grasps shown in an upper shadow and a lower shadow respectively. In the
upper shadow, the objects are at two different poses. The two grasps
associated with the two poses are identical in the local coordinate system
of the object. They are essentially the same grasp that represents the upper
node in the transit manifold.

Details of planning across multi-modal configuration spaces are shown in
Fig.\ref{multimodalmp}(c). Here, the object is a three-direction pipe (the
orange object in the figure). $\mathbf{P}^i$ is a standing pose shown in the
left part of Fig.\ref{multimodalmp}(b).
$\mathbf{P}^j$ is a resting pose shown in the right part of
Fig.\ref{multimodalmp}(b). The arm configurations in
$\mathcal{C}$($\mathbf{P}^i$) are related to the standing pose. The arm
configurations in $\mathcal{C}$($\mathbf{P}^i$) are related to the resting pose.

When doing sampling-based motion planning in a transfer space, a planner
will sample the space considering grasp configurations. 
An example is shown in Fig.\ref{multimodalmp}(c). There are two configuration
spaces $\mathcal{C}(\mathbf{P}^i)$ and $\mathcal{C}(\mathbf{P}^j)$ in the figure. The two spaces are connected
to each other by a transit space  $\mathcal{T}$($\mathbf{P}^i, \mathbf{P}^j)$. A
sampling-based planner samples the configuration spaces considering grasps, and each of the two configuration spaces in the figure are
sampled into five nodes (see the yellow and purple points in the figure). Nodes
$n_1$, $n_2$, $n_3$, $n_4$, $n_5$ belong to $\mathcal{C}(\mathbf{P}^i)$ . Nodes
$n_6$, $n_7$, $n_8$, $n_9$, $n_{10}$ belong to $\mathcal{C}(\mathbf{P}^j)$ .
In these sampled nodes, $n_1$, $n_2$, $n_3$ are further illustrated using green
hand under the $\mathcal{C}(\mathbf{P}^i)$ shadow. They belong to the transfer
space. $n_4$, $n_5$ are also further illustrated using green hand under the
$\mathcal{T}$($\mathbf{P}^i, \mathbf{P}^j)$ shadow. They belong to the transit space. $n_4$, $n_5$,
$n_6$, and $n_9$, $n_{10}$ are illustrated in the same way under the
$\mathcal{C}(\mathbf{P}^j)$ shadow and $\mathcal{T}$($\mathbf{P}^i, \mathbf{P}^j)$ shadow respectively.

The sampled nodes in $\mathcal{C}$($\mathbf{P}^i$), 
$\mathcal{C}$($\mathbf{P}^j$), and $\mathcal{T}$($\mathbf{P}^i$,
$\mathbf{P}^j$), are connected into a graph considering collisions and
feasibility of IK. The graph is a preliminary form of the regrasp graph. We will
further organize the preliminary graph considering both arms as well as handover
states in following sections to perform single-arm and dual-arm planning.

\section{Regrasp Graph}

This section presents the details of the regrasp graph, including automatic
grasp planning, selection of object poses, organizations of nodes,
single-arm regrasp graphs, and dual-arm extensions, etc.

\subsection{Grasps and object poses}

The fundamental elements of a regrasp graph are grasp configurations. In our
planner, the grasp configurations of an object are planned considering force
closure using self-developed grasp planners.

The object poses that modalize the configuration spaces are defined by
the stable placements of an object on a table. The stable
placements are planned considering the projection of $com$ (center of mass) on
a supporting area using a self-developed placement planner.
Especially we limit the number of object poses by discretizing the positions of
a stable placement and its rotation around the vertical axis. A
discretized placement, together with its accessible grasps and the collision-free and
IK-feasible configurations of a robot, defines a configuration space
$\mathcal{C}_i$. The accessible grasps will be the nodes of a
graph in $\mathcal{C}_i$.

\begin{figure*}[!htbp]
	\centering
	\includegraphics[width=\textwidth]{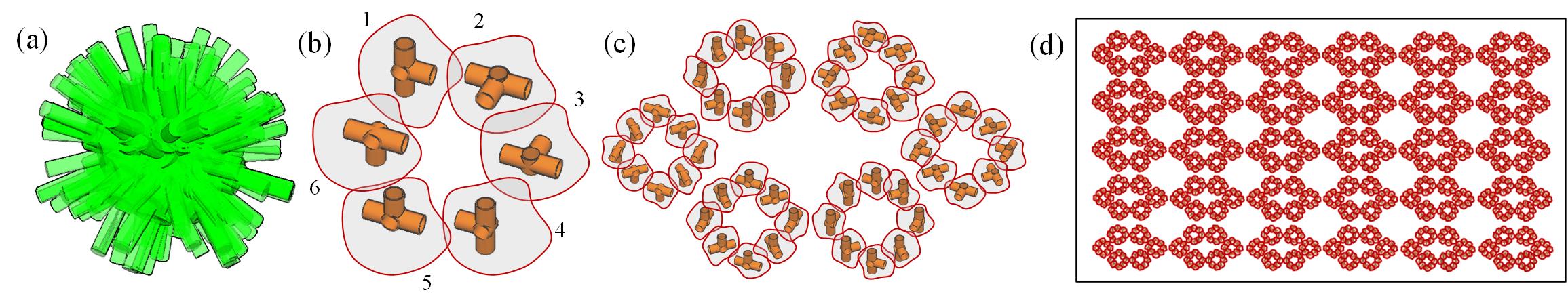}
	\caption{(a) The automatically planned grasps of a pipe model.
	(b) The stable placements (and the configuration spaces modalized by the
	placements) of the pipe on a table without considering rotation around vertical
	axis. There are 6 stable placements and 6 configuration spaces modalized by
	them. (c) The stable placements after considering rotation around vertical
	axis. The rotation angle is discretized into eight values. Thus, the poses in
	(b) further spawn 6$\times$8=48 stable placements, corresponding to 48
	configuration spaces. There are 6 groups. Objects belonging to the same group
	are at the same stable placement with different orientations.
	(d) The stable placements at discretized positions on a table.}
	\label{graphsrot}
\end{figure*}

Fig.\ref{graphsrot} shows the modalization of configuration spaces
using the three-direction pipe model. The planned grasps are shown in
Fig.\ref{graphsrot}(a). The stable placements of the pipe on a table without
considering variation in positions and rotations are shown in
Fig.\ref{graphsrot}(b)\footnote{Note that we do not consider symmetry in
computing these poses. For example, pose 2 and pose 3 are laying on an
upper face and bottom face respectively. They are symmetrically the same, but are treated as
different stable placements.}. Fig.\ref{graphsrot}(c) shows stable placements
considering rotation around the vertical axis. In the figure, the rotation
is descritized by \rev{$\frac{\pi}{2}$} intervals for clear visualization and easy
computation. Each placement in Fig.\ref{graphsrot}(b) therefore spawns
eight poses, and the total number of poses reaches to 48.


\begin{figure*}[!htbp]
	\centering
	\includegraphics[width=\textwidth]{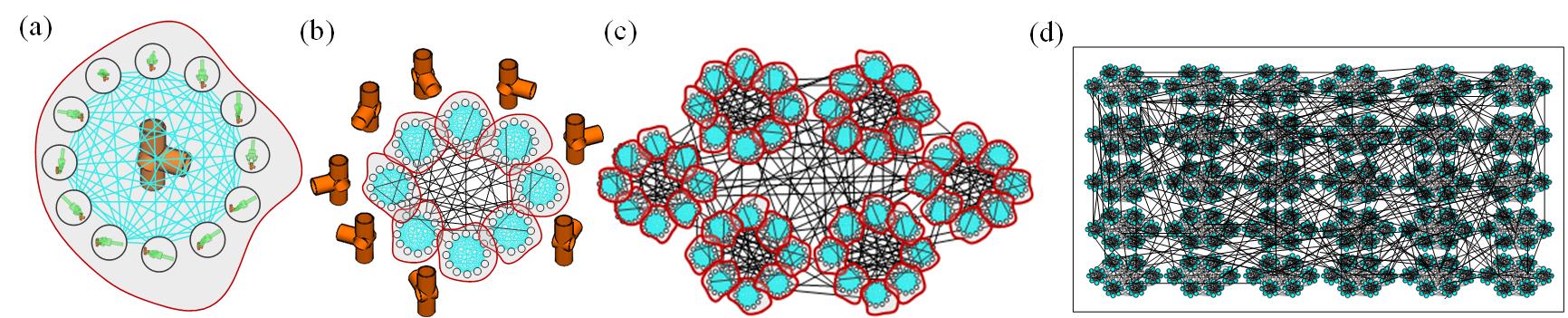}
	\caption{(a) The samples and roadmap
	built in one configuration space. Each circle indicates one sample. The
	samples are connected by transfer edges (cyan segments). (b) The roadmap
	built across the configuration spaces of one stable placement at 8
	different orientation. There are totally 8 configuration spaces. Transit edges
	(cyan segments) connect the samples belonging to the sample configuration
	space. Transfer edges (black segments) connect the samples belonging to
	different configuration spaces.
	(c) The roadmap built across the configuration spaces modalized by
	all stable placements at the 8 discretized orientation. (d) The samples and
	roadmap built across the configuration spaces at discretized positions all over
	a table.}
	\label{graphsgrasp}
\end{figure*}

Fig.\ref{graphsrot}(d) further shows the results after considering the
variation in positions. The figure is a top view where the black frame indicates
a table. Like rotation, the positions on the table are sampled by grids for
clear visualization and easy computation.
5$\times$6 grids are used for sampling and in total, the positions on the table are
discretized into 5$\times$6 locations. The 48 poses in Fig.\ref{graphsrot}(c) are further moved to these locations. In
total, after considering the positions on a table, the number of poses is 1440.
Each of the 1440 poses identifies one configuration space, and the
multi-modal planner shall plan across 1440 configuration spaces to find a preparatory
manipulation sequence for the three-direction pipe.

Note that the discretization of rotations and positions significantly
impair the scale of the regrasp graph. More rotations and positions increase
manipulability, as well as increase the scale of a regrasp graph. It adds heavy
work load to building the regrasp graph.
The scale of a regrasp graph is asymptotic to $n_g\times n_r \times n_p$ where $n_g$ is the number of
planned grasps, $n_r$ is the number of discretized rotation, $n_p$ is the
number of discretized positions. The number of planned stable poses is small
compared to these values and is therefore considered as a constant. Using
modern computers and databases, the planner may plan across as many as 10,000s
of nodes \cite{wan2017iros}.


\subsection{Single-arm regrasp graph}

To plan across the multiple configuration spaces, we sample nodes in each of
the configuration space and build a roadmap for
search by connecting the sampled node.
Fig.\ref{graphsgrasp}(a) shows a configuration space and the sampled nodes
(denoted by the circled grasps). The configuration space is modalized by a pose
of the object shown in the center.
Each sampled node is one collision-free and IK-feasible grasp of the object at
the pose. A node $\textbf{g}_k$ is
formally defined as a sample of $\mathcal{C}(\mathbf{P}^i)$, following equation
\rev{(\ref{posdim})}:
\begin{equation}
	\textbf{g}_k\in\mathcal{C}(\textnormal{\textbf{P}}^i)=\{d_g, \textbf{q},
	\textnormal{\textbf{T}}_{\textnormal{\textbf{P}}^i}\}
\end{equation}
The transfer edges $\textbf{E}_{tf}$ connecting the sampled nodes in
$\mathcal{C}(\mathbf{P}^i)$ are formally defined as:
\begin{equation}
    \rev{
	\textbf{E}_{tf}=
	\{(\textbf{g}_{k_0}, \textbf{g}_{k_1})|\textbf{g}_{k_0},\textbf{g}_{k_1}\in\mathcal{C}(\textnormal{\textbf{P}}^i)\wedge
	\textbf{q}^{(k_0)}\neq\textbf{q}^{(k_1)}\}}
	\label{etr}
\end{equation}
\rev{, where $\textbf{q}^{(k_0)}$ and $\textbf{q}^{(k_1)}$ are the $\textbf{q}$ components
of $\textbf{g}_{k_0}$ and $\textbf{g}_{k_1}$ respectively.}

In the case
of Fig.\ref{graphsgrasp}(a), there are 12 sampled grasps and consequently 12 nodes.
These nodes are connected into a graph by
transfer edges. The cyan edges in Fig.\ref{graphsgrasp}(a) indicate the transfer edges.
The graph is named a partial graph since it is part of the whole roadmap.

Fig.\ref{graphsgrasp}(b) further shows the result after considering rotation
around the vertical axis.
There are 8 configuration spaces and consequently 8 partial
regrasp graphs (shown in cyan color). Inside each partial regrasp
graph, the transfer edges connect sampled nodes. It is the partial roadmap
of a transfer space. Across the partial regrasp graphs, transit edges
(the black edges) connect the sampled nodes of the transfer spaces through
transit manifolds. The transit edges $\textbf{E}_{tt}$ are formally defined
as:
\begin{eqnarray}
    \rev{
	\textbf{E}_{tt}=\{(\textbf{g}_{k_0},
	\textbf{g}_{k_1})|\textbf{g}_{k_0}\in\mathcal{C}(\textnormal{\textbf{P}}^i)\wedge\textbf{g}_{k_1}
	\in\mathcal{C}(\textnormal{\textbf{P}}^j)\wedge \nonumber}\\
	\rev{
	\textnormal{\textbf{T}}_{\textnormal{\textbf{P}}^i}\cdot
	J(\textbf{q}^{({k_0})}) = \textnormal{\textbf{T}}_{\textnormal{\textbf{P}}^j}\cdot
	J(\textbf{q}^{({k_1})})\}}
	\label{ett}
\end{eqnarray}
\indent
Fig.\ref{graphsgrasp}(c) shows the regrasp graph after considering variation in
poses. There are 48 configuration spaces and the regrasp graph involves
48 partial regrasp graphs as well as the transit edges connecting them.
Fig.\ref{graphsgrasp}(d) further considers the different positions on a table.
In this case, there are 1440 configuration spaces and consequently 1440 partial
regrasp graphs. Together with the transit edges, the regrasp graph grows into a
big graph shown in Fig.\ref{graphsgrasp}(d). The graph includes 30 components
like Fig.\ref{graphsgrasp}(c). Each component corresponds to a different
position on a table. The additional edges in (c) and (d) follow the same
definition as equation (\ref{etr}), except that more object poses with varying
rotation around the vertical axis and different positions on a table are
further considered.

The regrasp graph in Fig.\ref{graphsgrasp}(d) is for a single arm. The grasps
represented by the nodes are only feasible to one arm. Thus, we name the graph
\textit{single-arm regrasp graph}, and use $\mathbf{G}_{x}$ to denote it.
\begin{equation}
	\rev{\mathbf{G}_{x}=\{\{\textbf{g}_k|\textbf{g}_k\in\mathcal{C}(\textnormal{\textbf{P}}^i)\},
	\textbf{E}_{tf},
	\textbf{E}_{tt}\}}
\end{equation}
$x$ could be replaced by $l$ or $r$.
$\mathbf{G}_{l}$ denotes the \textit{single-arm regrasp graph} for a left
arm. $\mathbf{G}_{r}$ denotes the \textit{single-arm regrasp graph} for a
right arm.

\subsection{Dual-arm regrasp graph}

\begin{figure*}[!htbp]
	\centering
	\includegraphics[width=\textwidth]{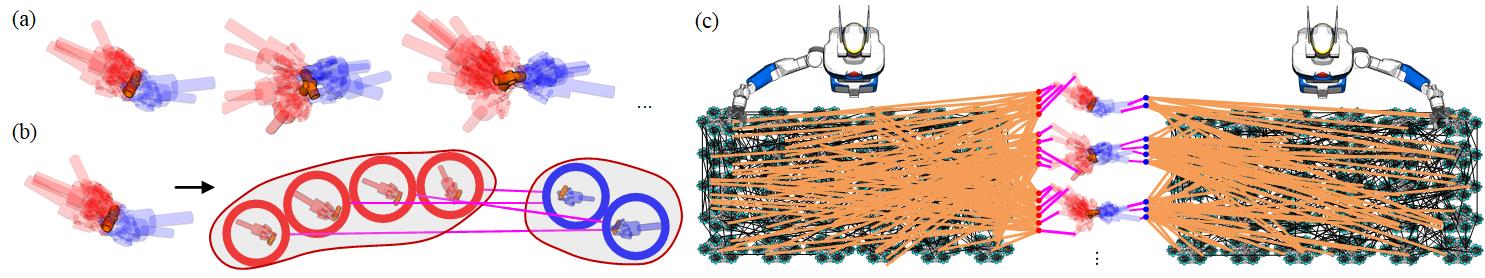}
	\caption{(a) The hand-over poses and the accessible grasps of
	two hands (red and blue). (b) One hand-over pose modalizes two configuration
	spaces. One is for the right hand. The other is for the left hand.
	(c) The \textit{dual-arm regrasp graph}.}
	\label{graphshandover}
\end{figure*}

The \textit{single-arm regrasp graph} $\mathbf{G}_{x}$ could be extended
to \textit{dual-arm regrasp graph} by building two regrasp graphs for two arms
and additionally including a \textit{hand-over regrasp graph}.

The hand-over graph is made by two modals identified by a hand-over pose of
the object. The nodes in the graph are the accessible grasps associated with the
hand-over pose. To extend to dual-arm regrasp, we build a \textit{single-arm
regrasp graph} for the right arm, and build another \textit{single-arm regrasp
graph} for the left arm. These two graphs are connected to the two modals in the
hand-over graph. The hand-over graph acts as a bridge for planning hand-over
motions.

See Fig.\ref{graphshandover} for details. First, the planner prepares
candidate hand-over poses to identify hand-over modals. The hand-over poses of
an object are discretized using 3D lattice and
icosphere. Each crossing point of a 3D lattice in the \rev{workspace} is used
as a candidate hand-over position. Each vector pointing to a vertex of an
icosphere centered at a position determined by the 3D lattice is used as a
hand-over orientation. Fig.\ref{graphshandover}(a) shows three hand-over poses.
These hand-over poses are at the same position. The position is determined
by a single crossing point of a 3D lattice. On the other hand, the hand-over
poses are at different orientations. The orientations are determined by three
vertices on \rev{an} icosphere centered at the aforementioned single crossing point.
The accessible grasps for two hands are shown in red and blue respectively.
Each of the hand-over poses modalizes two configuration spaces. One is for
the right hand, the other is for the left hand.

The first pose of
Fig.\ref{graphshandover}(a) corresponds to two configuration spaces (right-arm
configuration space and left-arm configuration space) shown in
Fig.\ref{graphshandover}(b).
The sampled nodes of the configuration spaces are shown by the grasps
in the red or blue circles. These sampled nodes are used for
hand-over. They are connected to each other to form a \textit{hand-over
regrasp graph}, considering the collisions between the two hands and the feasibility of
IK. 

We use $\textbf{P}^{i(o)}$ to denote a hand-over pose. Following equation
\rev{(\ref{posdim})}, the configuration spaces modalized by $\textbf{P}^{i(o)}$ is
defined as:
\begin{equation}
	\mathcal{C}(\mathbf{P}^{i(o)})^x = \{d_g, \textbf{q}^x,
	\textnormal{\textbf{T}}_{\textnormal{\textbf{P}}^{i(o)}}\}
\end{equation}
\indent
Here, $\textbf{q}^x$ \rev{could be} the joint angles of left arm ($\textbf{q}^l$) or right
arm ($\textbf{q}^r$). Accordingly, $\mathcal{C}(\mathbf{P}^{i(o)})^x$ is written
as $\mathcal{C}(\mathbf{P}^{i(o)})^l$ or $\mathcal{C}(\mathbf{P}^{i(o)})^r$.
\indent
The edges connecting two grasps associated with a hand-over pose are:
\begin{eqnarray}
    \rev{\textbf{E}_{tf}^{(o)}=\{(\textbf{g}_{k_0},
	\textbf{g}_{k_1})|\textbf{g}_{k_0}\in\mathcal{C}(\mathbf{P}^{i(o)})^l\wedge
	\textbf{g}_{k_1}\in\mathcal{C}(\mathbf{P}^{i(o)})^r\wedge}\nonumber\\
	\rev{M(\textbf{g}_{k_0})\cap M(\textbf{g}_{k_1})=\emptyset\}}
	\label{etrh}
\end{eqnarray}
Here, $M(\textbf{g}_{k_0})\cap M(\textbf{g}_{k_1})=\emptyset$ indicates the mesh
models of the gripper at the two configurations do not collide with each other. The object
poses at the end of the edges are the same, thus the edges are transfer edges.

The edges connecting the placements and hand-over poses are:
\begin{eqnarray}
    \rev{\textbf{E}_{tt}^{(o)}=\{(\textbf{g}_{k_0},
	\textbf{g}_{k_1})|\textbf{g}_{k_0}\in\mathcal{C}(\textnormal{\textbf{P}}^i)\wedge
	\textbf{g}_{k_1}\in\mathcal{C}(\textnormal{\textbf{P}}^{j(o)})\wedge}\nonumber\\
	\rev{\textnormal{\textbf{T}}_{\textnormal{\textbf{P}}^i}\cdot J(\textbf{q}^{(i)}) =
	\textnormal{\textbf{T}}_{{\textnormal{\textbf{P}}}^{j(o)}}\cdot
	J(\textbf{q}^{(j(o))})\}}
	\label{etth}
\end{eqnarray}
These edges are transit edges.

The \textit{hand-over regrasp graph} is denoted by $\mathbf{G}_{o}$ where
\begin{equation}
    \rev{
	\mathbf{G}_{o}=\{\{\textbf{g}_k|\textbf{g}_k\in\mathcal{C}(\mathbf{P}^{i(o)})^x\},
	\textbf{E}_{tf}^{(o)}\}}
\end{equation}
$\mathbf{G}_{o}$ is connected to the two
\textit{single-arm regrasp graphs} to build a \textit{dual-arm regrasp graph}
like Fig.\ref{graphshandover}(c). The dual-arm regrasp graph is denoted by
$\mathbf{G}$$_{(\mathbf{G}_{l}, \mathbf{G}_{o},
\mathbf{G}_{r})}$.
Here, $\mathbf{G}_{l}$ and $\mathbf{G}_{r}$ are the two \textit{single-arm
regrasp graphs} built for right and left arms. $\mathbf{G}_{l}$ and
$\mathbf{G}_{o}$, or \rev{$\mathbf{G}_{r}$} and $\mathbf{G}_{o}$, are connected to
each other by \rev{$\textbf{E}_{tf}^{(o)}$}.



\section{Regrasp Planning using Regrasp Graphs}

\subsection{Single-arm planning}

\begin{figure}[!htbp]
	\centering
	\includegraphics[width=.487\textwidth]{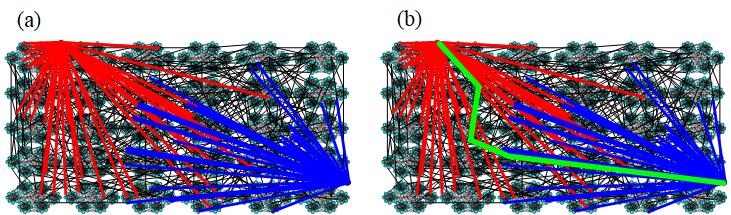}
	\caption{Single-arm planning. (a) The partial regrasp graphs
	of initial and goal poses are connected to the single-arm regrasp graph. The
	red segments are connected to the initial graph. The blue segments are
	connected to the goal graph. (b) A path found by searching from the initial to
	the goal.}
	\label{sarmsearch}
\end{figure}

Given the initial and goal poses of an object, the proposed planner
finds the accessible grasps of the two poses and generates the \textit{partial
regrasp graphs for the initial and goal poses}. $\mathbf{G}_i$ and $\mathbf{G}_g$ are used to denote the
\textit{partial regrasp graphs for the initial and goal poses}. The planner connects $\mathbf{G}_i$ and $\mathbf{G}_g$
to a \textit{single-arm regrasp graph} $\mathbf{G}_{x}$, and searches the
connected graph $\textbf{G}_{(\textbf{G}_x, \textbf{G}_i, \textbf{G}_g)}$ to
find preparatory manipulation sequences for a single arm.
Fig.\ref{sarmsearch} shows an example. The \textit{partial regarsp graphs for
the initial and goal poses} $\textbf{G}_i$ and $\textbf{G}_g$ are similar to the
one shown in Fig.\ref{graphsgrasp}(a), as the initial and goal poses are assumed to be a stable placement on a table. 
$\textbf{G}_i$ and $\textbf{G}_g$ are connected to a single-arm regrasp graph to build
$\mathbf{G}$$_{(\mathbf{G}_{x}, \mathbf{G}_{i}, \mathbf{G}_{g})}$
in Fig.\ref{sarmsearch}(a). Here, $\textbf{G}_i$ and $\textbf{G}_g$ can be
hardly seen since they are overwhelmed by the red and blue segments (they are
near the spots with heavy red and blue colors).
The planner searches the shortest path from $\textbf{G}_i$ to
$\textbf{G}_g$ to plan single-arm preparatory manipulation sequences for the task.
The initial and goal grasps could be any node on $\textbf{G}_i$ and
$\textbf{G}_g$.
The green path in
Fig.\ref{sarmsearch}(b) shows one path found by the planner.

The output of the single-arm planning is a sequence of collision-free and
IK-feasible grasps, and the result of single-arm planning is essentially a
sequence of grasps and robot configurations. The motion between the
configurations in the sequence is not necessarily feasible \rev{due to obstacles,
unavailable IKs, and some other task assumptions like the end-effector
constraints \cite{kingston18}}, especially in complicated \rev{workspaces}. 
\rev{On-line sampling-based motion planning} is therefore employed in
a later step to further generate detailed arm motion. \rev{Dynamic Domain RRT (DD-RRT) \cite{yershova2005dynamic}
is used to generate samples and plan paths}. \rev{DD-RRT is used because} we assume 
the \rev{workspace} \rev{is not crowded with obstacles} and \rev{the configuration space does not}
have many narrow passages. \rev{DD-RRT searches more in the open space and 
finds paths quickly under these assumptions}. \rev{For configuration spaces
connected by narrow passages, DD-RRT is a bit slow. 
Methods like \cite{Wan2015arso} are preferable.}

The sequence planning and motion planning are performed sequentially and
repeatedly in a loop shown in Alg.\ref{alg}. If motion planning fails, the
blocked edge will be removed from the regrasp graph and the single-arm regrasp
planner replans \rev{(the highlighted section of Alg.\ref{alg})}.

\begin{algorithm}[!htbp]
  \SetKwData{Null}{null}
  \SetKwFunction{search}{search}
  \SetKwFunction{DDRRT}{DDRRT}
  \SetKwFunction{removeblockededge}{removeblockededge}
  \DontPrintSemicolon
  \KwData{The regrasp graph $\mathcal{G}_{(\mathbf{G}_{x}, \mathbf{G}_{i}, \mathbf{G}_{g})}$}
  \KwResult{Motion for preparatory manipulation}
  \Begin {
  	\While{True}{
	  	path = \search{$\mathcal{G}_{(\mathbf{G}_{x}, \mathbf{G}_{i},
	  	\mathbf{G}_{g})}$}\\
	  	\If{\textnormal{path} is None}{
	  		\Return{None}
	  	}
	  	motion = \DDRRT{\textnormal{path}}\\
	  	\HiLi
	  	\If{\textnormal{motion} is Blocked}{
	  	    \HiLi
	  		\removeblockededge{$\mathcal{G}_{(\mathbf{G}_{x}, \mathbf{G}_{i},
	  	\mathbf{G}_{g})}$, \textnormal{path}}\\
	  	    \HiLi
	  		\textit{Continue}
	  	}
		\Return{\textnormal{motion}}
	  }
  }
  \caption{Incorporating Motion Planning}
  \label{alg}
\end{algorithm}

\subsection{Dual-arm planning}

Dual-arm planning is essentially the same as the single-arm planning, except
that $\textbf{G}_i$ and $\textbf{G}_g$ are respectively connected to
the $\textbf{G}_r$ and $\textbf{G}_l$ components of a \textit{dual-arm regrasp
graph}.

\begin{figure}[!htbp]
	\centering
	\includegraphics[width=.487\textwidth]{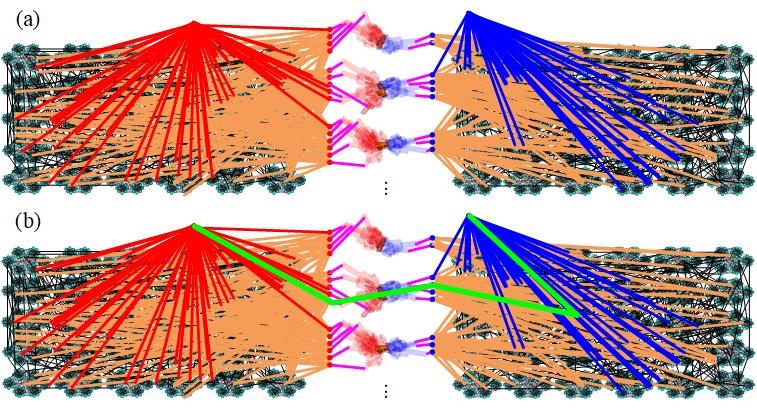}
	\caption{Dual-arm planning. (a) The partial regrasp graph $\textbf{G}_i$
	is connected to the right arm (red segments). The partial regrasp graph
	$\textbf{G}_g$ is connected to the left arm (blue segments). (b) A path found
	by searching from the initial to the goal. In this case, the hand-over
	direction is from right hand to left hand.}
	\label{darmsearch}
\end{figure}

Fig.\ref{darmsearch} shows an example.
Like the single-arm case, the planner searches the shortest
path from the initial regrasp graph to the goal regrasp graph, and performs
iterative motion planning to generate preparatory manipulation motion.
In the case of
Fig.\ref{darmsearch}, the hand-over is from right hand to left
hand since $\textbf{G}_i$ is connected to the
$\textbf{G}_r$ component (red segments) and
$\textbf{G}_g$ is connected to the $\textbf{G}_l$ component (blue segments).
The direction of the hand-over motion could the exchanged by swapping
the connection from $\textbf{G}_i\leftrightarrow\textbf{G}_r$,
$\textbf{G}_g\leftrightarrow\textbf{G}_l$ to
$\textbf{G}_i\leftrightarrow\textbf{G}_l$,
$\textbf{G}_g\leftrightarrow\textbf{G}_r$. Readers are recommended to refer
to \cite{Wan2016ral} for more details about single-arm and dual-arm planning.

\subsection{Automatically choosing the number of arms}

The most advantageous the new planner proposed in this work is it is able
to automatically choose the number of arms by building and searching a super regrasp graph which holds all connections among
$\textbf{G}_i$, $\textbf{G}_g$, and the components of
\textit{dual-arm regrasp graph}.
The connections include: 
\begin{itemize}
  \item Connection between $\textbf{G}_i$ and $\textbf{G}_r$.
  \item Connection between $\textbf{G}_g$ and $\textbf{G}_l$.
  \item Connection between $\textbf{G}_i$ and $\textbf{G}_l$.
  \item Connection between $\textbf{G}_g$ and $\textbf{G}_r$.
  \item Connection between $\textbf{G}_i$ and $\textbf{G}_g$.
\end{itemize}

Making up all the five connections leads to a super regrasp graph. Our planner
could search the super regrasp graph to find a preparatory manipulation
sequence which automatically determines whether to use single-arm regrasp,
dual-arm handover, or mixed single-arm regrasp and dual-arm handover, to perform
given tasks. The search is done using conventionally heuristic search
algorithm like A*\cite{hart1968formal}. The \rev{heuristic} we used is the length
of a path. Since the length of a path corresponds to the times of
transfer and transit changes or the times of regrasps,
our planner may find a path that requires fewest
times of regrasp. Meanwhile, it automatically determines if the path
spans the graphs of a single arm or both arms, hence automatically
determines the necessary number of arms. The shortest path
heuristic is also reasonable in that fewer times of regrasp reduce the
propagation of noises, making the preparatory manipulation more robust
\footnote{Note that there might be multiple solutions that require the same
times of regrasp. We do not further differentiate them in this work.
However, practitioners may choose their own strategy to rank these solutions. E.g. One may give high priority to human-like poses.}.

\begin{figure}[!htbp]
	\centering
	\includegraphics[width=.487\textwidth]{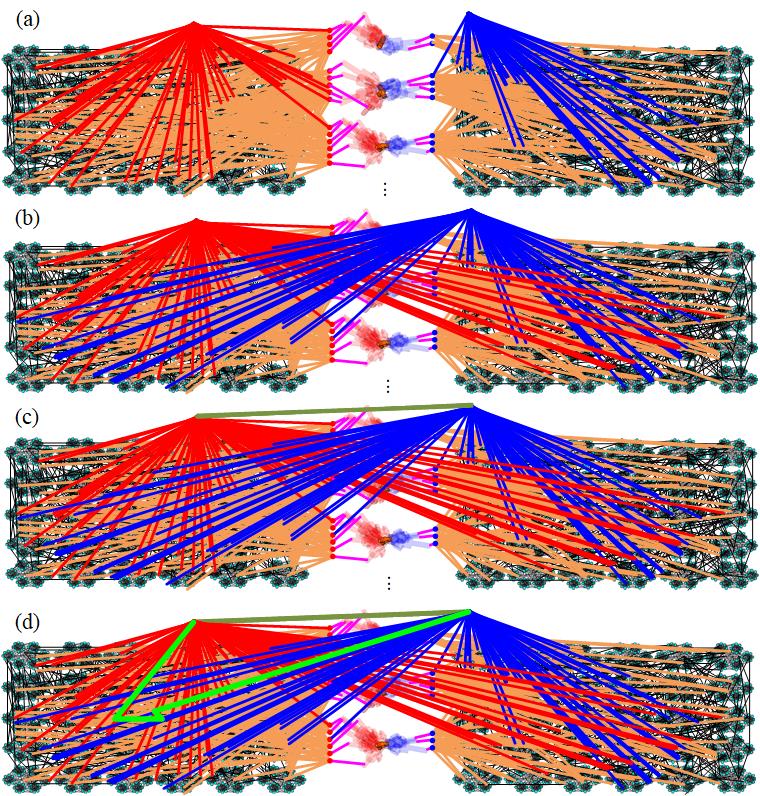}
	\caption{Mixed single and dual arm planning. (a) The partial regrasp graph
	$\textbf{G}_i$ is connected to the right arm (red segments). The partial regrasp graph
	$\textbf{G}_g$ is connected to the left arm (blue segments). (b) The partial regrasp graph
	$\textbf{G}_i$ is additionally connected to the left arm (additional red
	segments).
	The partial regrasp graph $\textbf{G}_g$ is additionally connected to the right
	arm (additional blue segments). (c) $\textbf{G}_i$ is connected to
	$\textbf{G}_g$ (brown segments). (d) A path found
	by searching from the initial to the goal. In this case, the planer
	determines to only use the right arm.}
	\label{aarmsearch}
\end{figure}

Fig.\ref{aarmsearch} shows an example.
In Fig.\ref{aarmsearch}(a), $\textbf{G}_i$ is connected to
$\textbf{G}_r$ (red segments) and $\textbf{G}_g$ is connected to $\textbf{G}_l$
(blue segments). In Fig.\ref{aarmsearch}(b), $\textbf{G}_i$ is additionally
connected to $\textbf{G}_l$ (red segments) and $\textbf{G}_g$ is additionally
connected to $\textbf{G}_r$ (blue segments).
In Fig.\ref{aarmsearch}(c), $\textbf{G}_i$ is connected to
$\textbf{G}_g$ (brown segments). Fig.\ref{aarmsearch}(c) is the super regrasp
graph after making up all the five connections.. A shortest path found by
searching the super graph is shown in Fig.\ref{aarmsearch}(d).

\section{Experiments and Analysis}

\subsection{Results using various robots}

\begin{figure*}[!htbp]
	\centering
	\includegraphics[width=\textwidth]{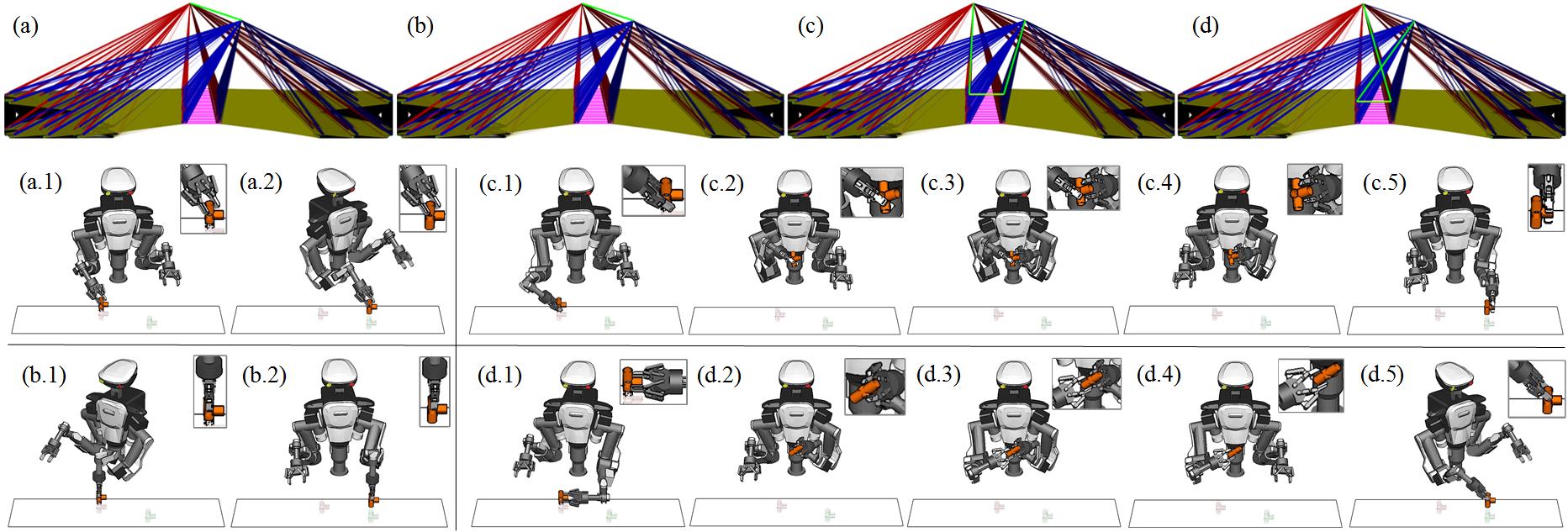}
	\caption{Kawada Nextage and a three-direction pipe --
	Example 1.
	The initial and goal poses of the object have the same orientation, but
	different positions. The planner suggests (a.1-a.2) or (b.1-b.2), which involves only one
	time of single-arm regrasp. (c.1-c.5) and (d.1-d.5) show another two solutions
	using one time of handover. They cost more than (a.1-a.2) and (b.1-b.2), and
	are consequently not suggested.}
	\label{result0}
\end{figure*}

\begin{figure*}[!htbp]
	\centering
	\includegraphics[width=\textwidth]{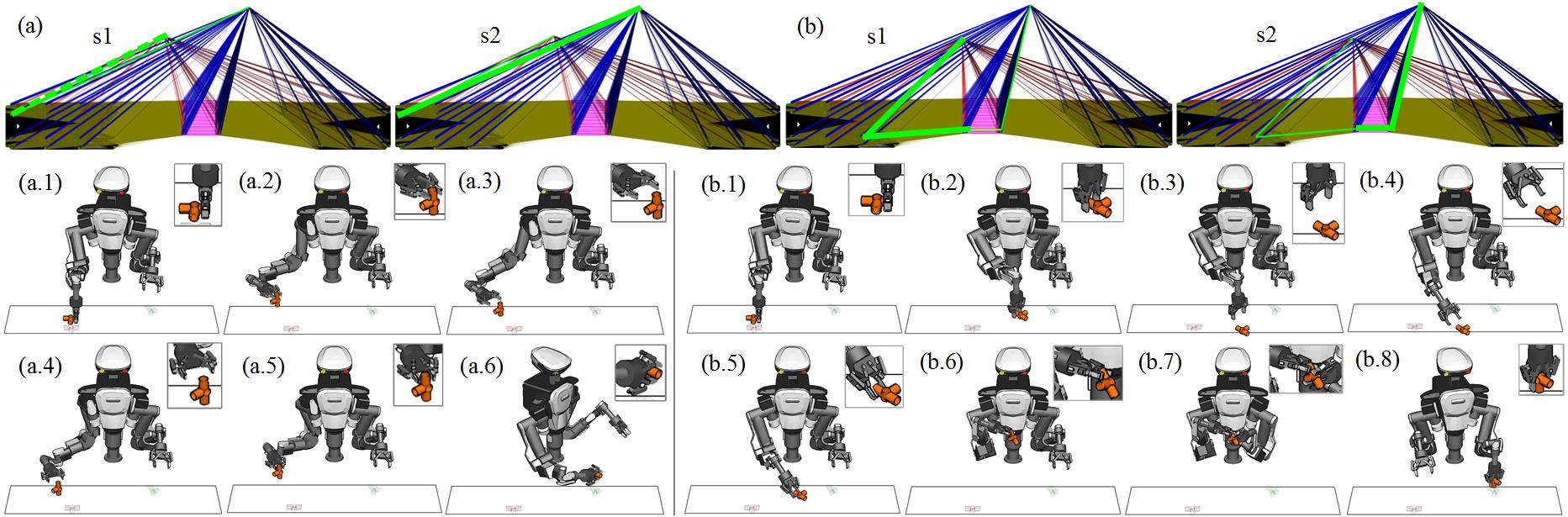}
	\caption{Kawada Nextage and a three-direction pipe --
	Example 2. In this case, the initial and goal poses are different in both
	positions and orientations. The robot could either use one time of single-arm
	regrasp, or one time of single-arm regrasp plus one time of hand over, to
	perform the task. The planner suggests (a.1-a.6) as it is shorter and
	consumes less energy.}
	\label{result1}
\end{figure*}

\begin{figure}[!htbp]
    \centering
    \includegraphics[width=.43\textwidth]{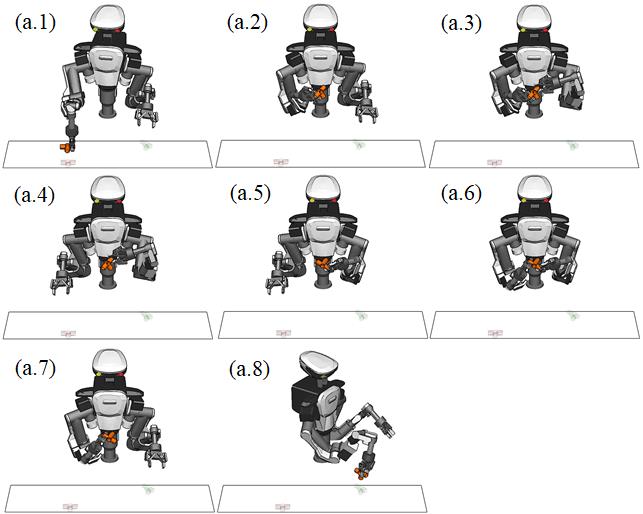}
    \caption{A special case where the planner suggested two times of handover
    to perform a task. (a.3, a.4): Hand over to left arm; (a.5): Reorient the
    object; (a.6, a.7): Hand over to right arm.}
    \label{resultadd}
\end{figure}

\begin{figure*}[!htbp]
    \centering
    \includegraphics[width=\textwidth]{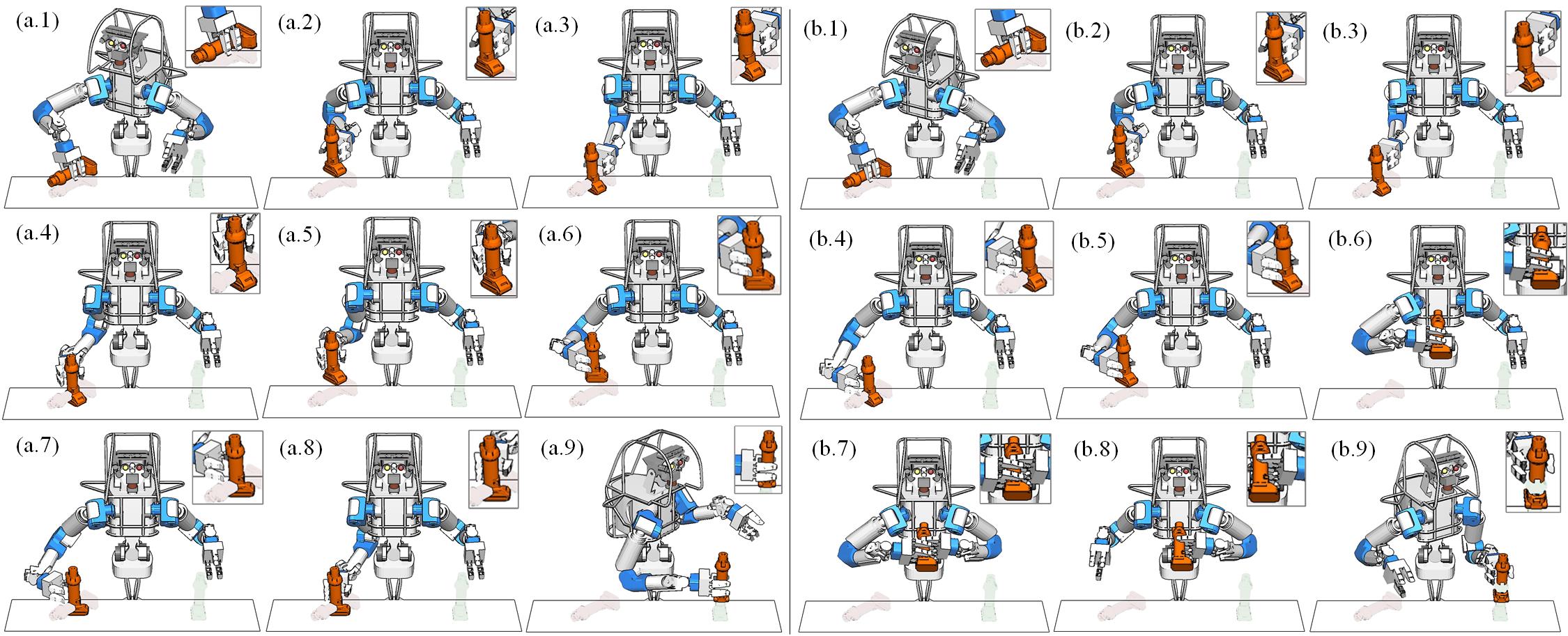}
    \caption{Using an HRP5P robot to reorient an electric drill. The robot
    could either use two times of single-arm regrasp (a.1-a.9) or one time of
    single-arm regrasp plus one time of handover (b.1-b.9) to perform the task.
    (a.1-a.9) is preferred.}
    \label{result2}
\end{figure*}

\begin{figure*}[!htbp]
    \centering
    \includegraphics[width=\textwidth]{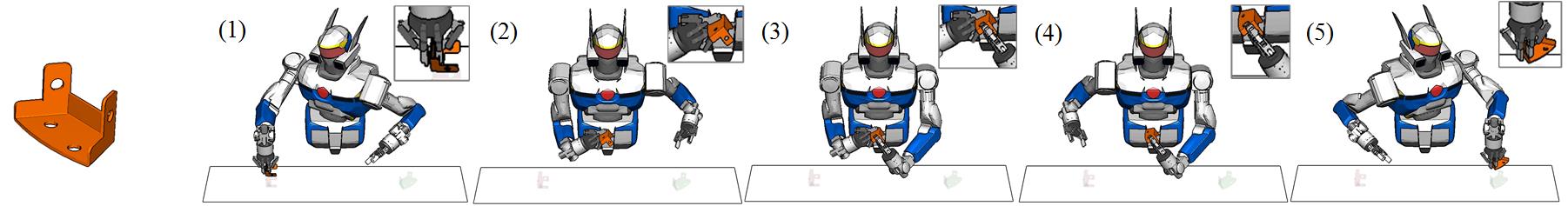}
    \caption{Using a HRP2Kai robot to reorient a workpiece. The only solution is
    to use handover shown in (1-5).}
    \label{result3}
\end{figure*}

The planner is tested using various robots including Nextage, HRP5P, and
HRP2Kai. Results of the planner and the correspondent regrasp graphs are shown
in Fig.\ref{result0}. The
task is to move objects from initial poses marked in
transparent red color to goal poses marked in transparent green color.
The results in Fig.\ref{result0}(a-d) perform a simple
pick-and-place task.
The pose of the object is not changed. The robot is Kawada Nextage.
Four sequences to finish the task are shown in Fig.\ref{result0}(a.1-a.2),
Fig.\ref{result0}(b.1-b.2), (c.1-c.4), and (d.1-d.4), respectively. The
regrasp graphs are shown in Fig.
\ref{result0}(a), (b), (c), and (d). Fig.\ref{result0}(a) and
(a.1-a.4) are the regrasp graphs and planned results by using the right arm.
Fig.\ref{result0}(b) and (b.1-b.4) are the regrasp graphs and planned results by
using the left arm. Fig.\ref{result0}(c) and (c.1-c.4) are the regrasp graph and
results by using right-to-left handover. Fig.\ref{result0}(d) and (d.1-d.4) are
the regrasp graph and results by using left-to-right handover. In this case, the
planner suggests using the shorter paths (a.1-a.2) or (b.1-b.2) to perform the
task, which involves only one time of single-arm regrasp. (c.1-c.5) and
(d.1-d.5) involve one time of handover. They cost more than (a.1-a.2) and
(b.1-b.2), and are consequently not suggested.

The results in Fig.\ref{result1}(a-b) use the same robot and object, but
the initial and goal poses of the object are changed. In this
case, the robot could either use two times of single-arm regrasp (a.1-a.6) or one time of
single-arm regrasp plus one time of handover (b.1-b.8) to perform the task.
The regrasp graph and path corresponding to (a.1-a.6) is shown in (a), where
the thick segment in (a)-s1 implies the pick-and-place sequence shown in
(a.1-a.3), the thick segment in (a)-s2 implies a second pick-and-place sequence
shown in (a.4-a.6).
The regrasp graph and path corresponding to (b.1-b.8) are shown in (b), where
the thick segments in (b)-s1 correspond the pick-and-place sequence shown in
(b.1-b.4), the thick segments in (a)-s2 correspond the handover sequence shown
in (b.5-b.8).
The planner suggests using the shorter path (a.1-a.6) to perform the task.
The results in Fig.\ref{resultadd} show a special case where the robot picks
up an object using its right arm, hands it over to uses its left arm to re-orient the
object, and hands it back to the right arm. The planner suggested two times of
handover to finish the task.
The results in Fig.\ref{result2}(a-b) show the result of reorienting an
electric drill using an HRP5P robot. In this case, The robot
could either use two times of single-arm regrasp (a.1-a.9) or one time of
single-arm regrasp plus one time of handover (b.1-b.9) to perform the task. The
planner outputs (a.1-a.9). The task cannot be done by simple pick-and-place or
direct handover.
The results in Fig.\ref{result3} show the result of reorienting a
workpiece using an HRP2Kai robot. In this case, the only solution is
to use handover shown in (1-5). The robot cannot perform the task using a
single arm.

\begin{figure}[!htbp]
    \centering
    \includegraphics[width=.47\textwidth]{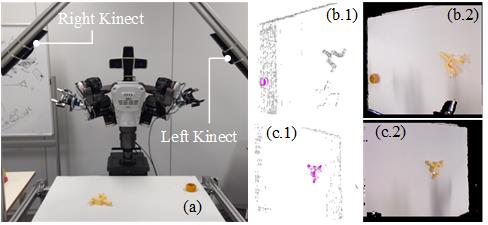}
    \caption{(a) The two external Kinect V1 cameras used for pose detection.
    (b) A matching result by using the right Kinect. (c) A matching result
    by using the left Kinect.}
    \label{visionsys}
\end{figure}

\begin{figure*}[!htbp]
    \centering
    \includegraphics[width=\textwidth]{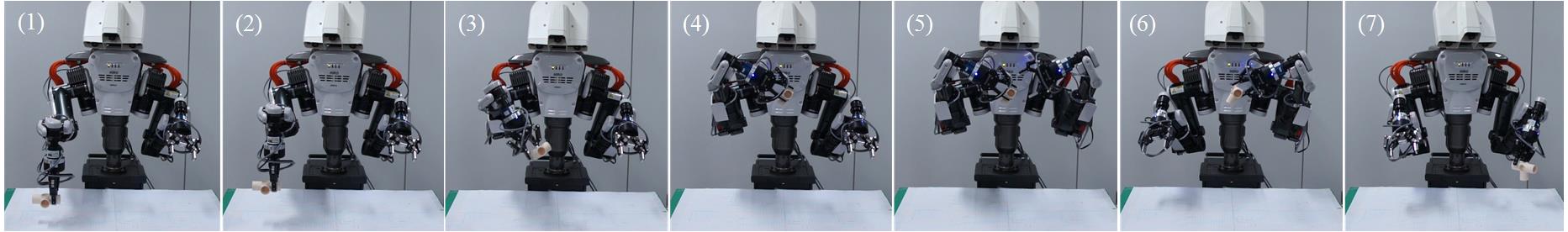}
    \caption{Real-world experiments using a Nextage robot. The executed sequence
    is Fig.\ref{result0}(d.1-d.5).}
    \label{result4}
\end{figure*}

\rev{
Real-world experiments are conducted on a Nextage robot.
In the real-world experiments, we used two external Kinect
V1 cameras and evaluated the pose of an object by minimizing the
distance between the 3D point clouds and a mesh model. The hardware
configurations and two matching results are shown in Fig.\ref{visionsys}. Each
detection takes about 1 second, which had been analyzed in detail in
\cite{Harada14}. There is no re-estimation during the regrasp.
The results
corresponding to Fig.\ref{result0}(d.1-d.5) are shown in Fig.\ref{result4}. The
robot automatically uses the shortest path (either single-arm, dual-arm, or
mixed single and dual arm regrasp) to finish the task.}

\begin{table*}[!htbp]
\centering
\renewcommand{\arraystretch}{1.2}
\caption{\label{dstc}Time cost}
\resizebox{0.99\linewidth}{!}{%
\begin{threeparttable}
\begin{tabular}{cccccccccccccccc}
\toprule
 \multirow{ 2}{*}{Robot} & \multirow{
 2}{*}{DoFs} &&
 \multicolumn{4}{c}{Online cost ($s$)} && \multicolumn{6}{c}{\gr{Offline cost
 ($s$)}} && \multirow{ 2}{*}{\# paths}\\
 \cmidrule{4-7} \cmidrule{9-14}
 &  && IK,CD($\textbf{G}_i$) & IK,CD($\textbf{G}_g$) & SG & $t$ &&
 \gr{gp} & \gr{pp} & \gr{IK,CD($\textbf{G}_x$)} & \gr{hp} &
 \gr{IK,CD($\textbf{G}_o$)} & \gr{$t$} &&
 \\
 \midrule
 Nextage & 12 && 8.00 & 13.72 & 2.29 & 24.01 && \gr{12.40} & \gr{7.21} &
 \gr{1702.11} & \gr{32.59} & \gr{1490.01} & \gr{3244.32} && 13.50\\
 HRP5P & 18 && 20.23 & 38.40 & 2.07 & 60.70 && \gr{10.22} & \gr{12.73} &
 \gr{3397.10} & \gr{54.29} & \gr{3670.27} & \gr{7177.20} && 6.48\\
 HRP2Kai & 14 && 10.65 & 19.88 & 1.44 & 31.97 && \gr{18.35} & \gr{11.36} &
 \gr{2552.51} & \gr{26.18} & \gr{1901.75} & \gr{4510.05} && 2.22\\
\bottomrule
\end{tabular}
\begin{tablenotes}[para,flushleft]
The online cost and number of paths are the average value of 10 executions.
All time costs are measured in seconds. Abbreviations: IK - Inverse
Kinematics; CD - Collision Detection; SG - Super Graph; gp - grasp planning;
pp - placement planning; hp - handover planning.
\end{tablenotes}
\end{threeparttable}}
\end{table*}

One thing to note is we build the regrasp graph with a large number of
nodes. It is redundant and offers many choices with equal times of regrasp. A robot may
switch to a different choice if the object is not well recognized by the vision
system. The visual system assumes a single object without surrounding obstacles.
The detection is robust under this assumption. We didn't spot a failure caused
by the vision system during the planning.

\subsection{Time cost and performance}


\subsubsection{\rev{Time cost}}
The planner includes an offline part and an online part. 
\rev{Both of them are computed automatically.} 
The offline part includes grasp planning, placement planning, handover planning, and
pre-building of partial roadmaps.  \rev{These computations are independent of
specific environments since they are irrelevant to obstacles. They can be built once and reused later.}
The online part builds supper roadmaps and performs graph search. Super
roadmaps have to be built online since the initial and goal poses of objects\rev{,
as well as the size of tables (obstacles)} change from task to task. 
The associated grasps and the edges that connect
these grasps to the partial roadmaps must be computed in realtime. The planner
searches all possible paths by using single-arms, dual-arms, or their
combinations and outputs the shortest path as the result. The cost of the
algorithms, including both the online and offline part are shown in
Table \ref{dstc}.
The processor of our computer is Intel Xeon 2.8GHz. Its
graphics card is NVIDIA Quadro M3000M. The version of the programming
language is Python 2.7.11. As is shown in Table \ref{dstc}, the offline part may cost
hours, depending on the number of DoFs. The online part, in contrast,
could be done in a few seconds. 
The most time-consuming part of the online cost is the computation of IK
and CD. In the worst case, the cost reaches to 38.40$s$. In contrast, building super
graphs and searching the graph is very fast. The most time-consuming part of
the offline cost is still the computation of IK and CD. Grasp planning and
placement planning also cost more than ten seconds. Handover planning costs
several tens of seconds. The best way to lower time cost is to reduce the
number of automatically planned grasps. The time cost table shows that the
Nextage robot has much better kinematic design than the other two robots. 
It has fewer DoFs, but could finish the given tasks with more sequences 
(see the \# paths column), and within
less time (see the online cost column).

\subsubsection{Performance}
\begin{figure}[!htbp]
    \centering
    \includegraphics[width=.45\textwidth]{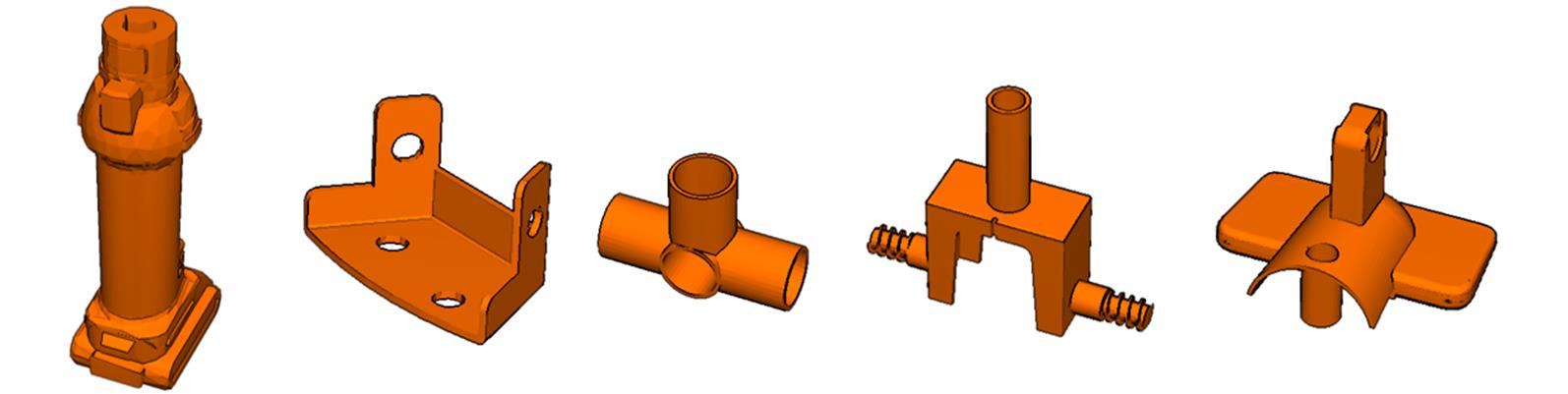}
    \caption{The objects used in the comparison experiments. From left to
    right: (1) An electric drill. (2) A metal work piece. (3) A three-way tube.
    (4) A connector. (5) A toy plane rear.}
    \label{fiveobj}
\end{figure}

Table \ref{pd} compares the performance of single-arm
regrasp planner, dual-arm regrasp planner, and the newly proposed mixed
planner using ten regrasp tasks and the objects shown in Fig.\ref{fiveobj}. The
initial and goal poses of these tasks are randomly generated. Compared to the
single-arm regrasp planner and dual-arm handover planner, the newly proposed
planner does preparatory manipulation planning using automatically determined
single and dual arms. It has higher success rates in reorienting. Also, the
planner is able to do both regrasp and handover, which significantly
improves the ability of humanoid robots.

\begin{table}[!htbp]
\centering
\renewcommand{\arraystretch}{1.2}
\caption{\label{pd}Performance of different planners}
\resizebox{0.99\linewidth}{!}{
\begin{threeparttable}
\begin{tabular}{cccc}
\toprule
  & Sgl-arm planner & Dual-arm planner & Auto planner
 \\
 \midrule
 $ir$-$gr$ & 7/10 & 8/10 & 10/10\\
 $il$-$gl$ & 5/10 & 8/10 & 9/10\\
 $ir$-$gl$ & $\times$ & 10/10 & 10/10\\
 $il$-$gr$ & $\times$ & 10/10 & 10/10\\
 \midrule
 regrasp & $\bigcirc$ & $\times$ & $\bigcirc$\\
 handover & $\times$ & $\bigcirc$ & $\bigcirc$\\
\bottomrule
\end{tabular}
\begin{tablenotes}[para,flushleft]
The success rates are based on ten times of executions. $ir$-$gr$ means initializing
from right arm and reaching to the goal using right arm. $ir$-$gl$ means
initializing from right arm and reaching to the goal using left arm. $il$-$gl$
and $il$-$gr$ follow similar definitions.
\end{tablenotes}
\end{threeparttable}}
\end{table}

\subsubsection{\rev{Discussions on more than two arms}} Extending the algorithms to
more than two arms is not difficult since the graphs are built incrementally.
More than two arms require including more $\mathbf{G}_x$ where $x$ indicates
the identifier of all arms, rather than the binary values $r$ and $l$ in
dual-arm planning. Accordingly, connections between the \textit{partial regrasp
graphs} and \textit{single-arm regrasp graphs} should include: (1) Connections between $\mathbf{G}_i$ and all $\mathbf{G}_x$. (2)
Connection between $\mathbf{G}_g$ and all $\mathbf{G}_x$. (3) Connection between $\textbf{G}_i$ and
$\textbf{G}_g$. Although incrementally building the connections between
$\mathbf{G}_i$, $\mathbf{G}_g$, and all other $\mathbf{G}_x$ do not
significantly increase computational cost, preparing the additional
$\mathbf{G}_x$s cost many extra offline resources. Accordingly to
Table \ref{dstc}, the offline time cost for a single $\mathbf{G}_x$ (6-DoF
Nextage) is around 1500$s$. The offline time cost of an n-arm 6-DoF robot will be as much as
1500n$s$. The high cost makes it impractical to extend the present algorithms to
multiple arms.

\section{Conclusions and Future work}

In this paper, we presented a preparatory manipulation planner to plan a
sequence of manipulation motion that prepared an object into expected poses. The planner
was demonstrated by various robots using both simulation and real-world
experiments. The results showed that the planner could plan single-arm,
dual-arm, as well as mixed single and dual arm preparatory manipulation sequences. It could
automatically determine the employment of single and dual arms by minimizing
the number of regrasp and handover. The planner is a \rev{high-level} component and
is applicable to various humanoid robots like HRP2Kai, Nextage, etc.

\rev{There are several future directions. First, as} discussed in the
experimental section, the present algorithms are not suitable \rev{for} more than
two arms due to the high offline computational cost. The present
algorithms plan over both arms in a centralized way, which we believe is not
applicable to more than two arms since few creatures in nature control more than
two dexterous arms using a central brain. Thus, we will explore decentralized
methods and study planning across multiple industrial robots (more than two
arms) or human-robot coordination using a
combination of the proposed algorithms and decentralized approaches.
\rev{Second,} the current visual detection system assumes a single object without
surrounding obstacles. Although the assumption eases visual estimation, in more
complicated environments where the object is obstructed and can hardly be
recognized from one view, we may use hand mounted cameras and active visual
perception to perform mixed view and preparatory manipulation planning.
\rev{Third, the placement planning algorithm is specially designed for tables. It would
be interesting and promising to develop general stable state planners that is
applicable to any supporting surfaces and unstructured environments. 
We have done some preliminary studies on it,
and will explore more in the future.} \rev{Moreover, changing environments, 
especially the environments where
human works together with the robot and has to be treated as a moving obstacle,
also deserves exploration. The current DD-RRT algorithm and regrasp replanning have
to be adapted to determined good timing for reactions.}

\bibliographystyle{ieeetr}
\bibliography{references}

\end{document}